\documentclass{article}

\usepackage{microtype}
\usepackage{graphicx}
\usepackage{subfigure}
\usepackage{booktabs} 

\usepackage{hyperref}

\usepackage[preprint]{icml2026}

\usepackage{amsmath}
\usepackage{amssymb}
\usepackage{mathtools}
\usepackage{amsthm}

\usepackage[capitalize,noabbrev]{cleveref}

\theoremstyle{plain}

\theoremstyle{definition}

\theoremstyle{remark}

\usepackage[textsize=tiny]{todonotes}
\usepackage{enumitem}
\usepackage{booktabs}
\usepackage{multirow}
\icmltitlerunning{Improving LLM Reasoning with On-Policy Thinking Language Selection}

\begin{document}

\twocolumn[
\icmltitle{ExpLang: Improved Exploration and Exploitation in \\LLM Reasoning with On-Policy Thinking Language Selection}




\begin{icmlauthorlist}
\icmlauthor{Changjiang Gao}{nju,cityu}
\icmlauthor{Zixian Huang}{ailab}
\icmlauthor{Kaichen Yang}{ailab,dlut}
\icmlauthor{Jiajun Chen}{nju}
\icmlauthor{Jixing Li}{cityu}
\icmlauthor{Shujian Huang}{nju}
\end{icmlauthorlist}

\icmlaffiliation{nju}{National Key Laboratory for Novel Software Technology, Nanjing University, Jiangsu, China}
\icmlaffiliation{ailab}{Shanghai Artificial Intelligence Laboratory, Shanghai, China}
\icmlaffiliation{cityu}{Department of Linguistics and Translation, City University of Hong Kong, Hong Kong, China}
\icmlaffiliation{dlut}{School of Mathematical Sciences, Dalian University of Technology, Liaoning, China}

\icmlcorrespondingauthor{Shujian Huang}{huangsj@nju.edu.cn}

\icmlkeywords{Machine Learning, ICML}

\vskip 0.3in
]



\printAffiliationsAndNotice{}  

\begin{abstract}
Current large reasoning models (LRMs) have shown strong ability on challenging tasks after reinforcement learning (RL) based post-training. However, previous work mainly focuses on English reasoning in expectation of the strongest performance, despite the demonstrated potential advantage of multilingual thinking, as well as the requirement for native thinking traces by global users. In this paper, we propose ExpLang, a novel LLM post-training pipeline that enables on-policy thinking language selection to improve exploration and exploitation during RL with the use of multiple languages. The results show that our method steadily outperforms English-only training with the same training budget, while showing high thinking language compliance for both seen and unseen languages. Analysis shows that, by enabling on-policy thinking language selection as an action during RL, ExpLang effectively extends the RL exploration space with diversified language preference and improves the RL exploitation outcome with leveraged non-English advantage. The method is orthogonal to most RL algorithms and opens up a new perspective on using multilinguality to improve LRMs\footnote{Our code and data is available at \url{https://github.com/RiverGao/ExpLang}}.
\end{abstract}

\begin{figure}[t]
    \centering
    \includegraphics[width=1\linewidth]{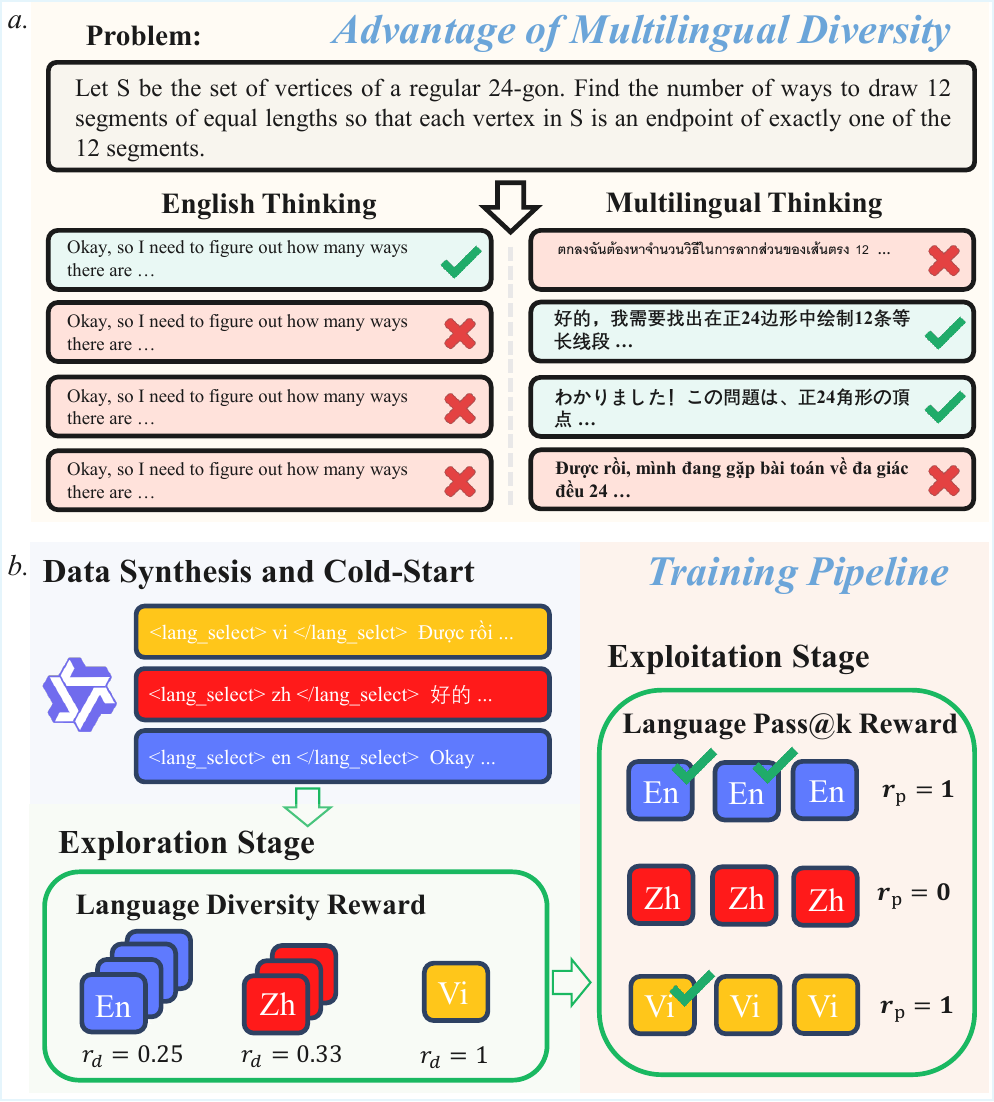}
    \vspace*{-15pt}
    \caption{Overview of our work. \textit{a.} Multilingual thinking shows larger diversity and higher Pass@$k$, suggesting better exploration space; \textit{b.} The ExpLang pipeline with on-policy thinking language selection to leverage the advantage.}
    \label{fig:overview}
\end{figure}

\section{Introduction}
\label{introduction}
Large language model (LLM) reasoning, as a form of the test-time scaling \cite{wuInferenceScalingLaws2024,muennighoffS1SimpleTesttime2025}, has been rapidly developing from Chain-of-Thoughts (CoT, \citealt{weiChainofThoughtPromptingElicits2022}) prompting to large reasoning models (LRMs) based on Reinforcement Learning with Verifiable Rewards (RLVR),  showing very strong performance on challenging tasks \cite{openaiOpenAIO1System2024,deepseek-aiDeepSeekR1IncentivizingReasoning2025}. 

However, current research mainly focuses on English reasoning, partially because the models tend to show their strongest performance in English, as a result of their English-dominant training data. For a long time, the multilinguality in LLM reasoning has been viewed as a ``curse'' that hinders performance \cite{etxanizMultilingualLanguageModels2024a}, or a service that caters to users with different language backgrounds \cite{zhangThinkNativelyUnlocking2025}. In contrast, some previous work point out that multilingual reasoning may have a potential benefit to overall performance \cite{chenLessDataLess2025,huangBenchMAXComprehensiveMultilingual2025,gaoCouldThinkingMultilingually2025,xuLanguageThoughtShapes2026}. For example, the model's Pass@$k$ score with multilingual CoTs is significantly higher than with English CoTs\cite{gaoCouldThinkingMultilingually2025}, suggesting that multilingual thinking extends the model's exploration space with higher expected return for RLVR. Thus, this work attempts to equip the model with self-controlled multilingual thinking during training and testing, leveraging its higher exploration and exploitation efficiency to enhance the overall reasoning ability.

In this work, we choose math reasoning as a representative task, and propose \textbf{ExpLang} (\textbf{Exp}lore and \textbf{Exp}loit with multiple \textbf{Lang}uages), a novel LLM post-training pipeline that encourages the model to explore reasoning with different languages and exploit the extended action space to improve reasoning performance, which are enabled by supervised cold-starting and RLVR with on-policy thinking language selection.
Figure \ref{fig:overview} shows the overview of our work.
Results show that our method:
\begin{itemize} [nosep,itemsep=2pt,leftmargin=0.2cm]
    \item Steadily outperforms English-only training with the same training budget, while showing significantly higher performance gain during RLVR;
    \item Shows near-perfect thinking language compliance under forced language selection for seen languages, generalizing well to unseen languages;
    \item Enables on-policy thinking language selection during RL, witnessing diversified language preference during the exploration stage and converged during the exploitation stage, benefiting the model with higher response diversity and leveraged non-English advantage.
\end{itemize}

To the best of our knowledge, this is the first work that advances in the above three aspects with a single training pipeline, which offers a novel perspective of covering and leveraging multilinguality in LLM reasoning with benefit in both language equality and model performance.

\section{Related Work}
\subsection{From Chain-of-Thoughts to RL-Enhanced Reasoning}
LLM reasoning is a common form of test-time scaling \cite{wuInferenceScalingLaws2024}, which originates from the chain-of-thought technique \cite{weiChainofThoughtPromptingElicits2022} that improves LLM performance with step-by-step answers. When combined with RL methods designed for LLMs \cite{schulmanProximalPolicyOptimization2017a,rafailovDirectPreferenceOptimization2024}, RL-empowered LRMs are proposed \cite{openaiOpenAIO1System2024,deepseek-aiDeepSeekR1IncentivizingReasoning2025,muennighoffS1SimpleTesttime2025}, which are unprecedentedly strong on challenging tasks such as math reasoning. Among them, one of the most influential work is GRPO \cite{shaoDeepSeekMathPushingLimits2024}, which effectively reduces the RL training cost by removing the value model, inspiring many new RL algorithms \cite{liuUnderstandingR1ZeroLikeTraining2025a,yuDAPOOpenSourceLLM2025b,zhengGroupSequencePolicy2025}.

\subsection{Multilingual Reasoning of LLMs}
Despite their success, LRM research highly focuses on English, where the role of multilinguality is largely neglected or debated. There are mainly four views regarding LRM multilinguality: (1) As a curse, which claims that introducing non-English languages in training or testing harms the models' performance \cite{qiWhenModelsReason2025,etxanizMultilingualLanguageModels2024a}; (2) As a user service, which underlines that LRMs should adjust their reasoning languages according to the user's need \cite{changWhenMultilingualityCurse2024,zhangThinkNativelyUnlocking2025,parkCrosslingualCollapseHow2025,liuConditionsCatastrophicForgetting2025}; (3) As a transferrable capability, which shows that LLMs are able to partly transfer their reasoning capabilities across languages under certain conditions \cite{zhangGettingMoreLess2024,huLargeLanguageModels2025,ranaldiMultilingualReasoningSelftraining2025};  (4) As a potential benefit, which demonstrates that non-English thinking can sometimes outperform English thinking, thus its advantages can be leveraged \cite{chenLessDataLess2025,gaoCouldThinkingMultilingually2025,huangBenchMAXComprehensiveMultilingual2025,xuLanguageThoughtShapes2026}. This paper proposes on-policy multilingual thinking as a useful method to not only improve LRM training, but also fostering user-compliance.
  
Current methods to perform multilingual thinking mainly falls in two types. The first is prefix-hacking, i.e. adding language-specific prefixes at the beginning of model generation \cite{yongCrosslingualReasoningTestTime2025,qiWhenModelsReason2025,xuLanguageThoughtShapes2026}, which is lightweight but unstable. The second is post-training, i.e. finetuning the model with multilingual thinking data \cite{laiMCoTMultilingualInstruction2024a,chaiXCoTCrosslingualInstruction2024b,sheMAPOAdvancingMultilingual2024c,baruaLongChainofThoughtReasoning2025,faisalAligningMultilingualReasoning2025,sonPushingMultilingualReasoning2025a,zhangThinkNativelyUnlocking2025}, which is stable but causes performance drop. This paper combines the two sides and proposes a stable and cost-efficient way of thinking language control.

\section{Building Blocks}
Starting from the idea of extending LLMs' exploration space with multilingual thinking to improve their mathematical reasoning, our method first equips the model with steady thinking language control, and then enable on-policy selection for multilingual exploration and exploitation in RLVR. The whole process is illustrated in Figure \ref{fig:overview}.


\subsection{High-Quality Multilingual Thinking Data}
Since current reasoning datasets are English-dominant, the first building step is to generate high-quality multilingual thinking data based on existing open-source models and datasets. Specifically, for a moderate-size open-source LLM $M_0$, we adopt its larger variant $\tilde{M}_0$ as the teacher to perform multilingual thinking on an English math training set, induced by 12 non-English prefixes, including 6 European and 6 Asian languages (see Figure \ref{fig:prefixes}). The language list acts as a meaningful representative, which can be extended with richer computational resource. Since the inference results contain incorrect answers and incompliant thinking languages, we identify the thinking languages with LangDetect \footnote{https://pypi.org/project/langdetect}and verify the answer correctness with HuggingFace Math-Verify \footnote{https://github.com/huggingface/Math-Verify}. The two-way filtration helps improve the quality of the multilingual thinking data.

To save inference cost, we first estimate the acceptance rate for each language with 100 samples (see Figure \ref{fig:prefixes}), and get approximately 500 filtered samples in each non-English language guided by these rates. To balance diversity and comparison within the dataset, inputs for each language are independently sampled from the original dataset. After that, we add a language selection tag for each sample in the form ``\texttt{<lang\_select>[lang]</lang\_select>}'' before the thinking tag, where \texttt{[lang]} is the language code such as \texttt{en}, \texttt{zh}, etc..

\subsection{Cold-Start of Thinking Language Selection}
The next building block is to align $M_0$ with the collected data to cold-start the thinking language behavior, for which SFT is a suitable choice. The cold-started model, $M_1$, is supposed to be highly compliant to the above mentioned language selection tags.

However, since $M_0$ is a post-trained model, it is necessary to prevent catastrophic forgetting during the finetuning process. In this regard, we use LoRA \cite{huLoRALowRankAdaptation2021d} SFT to reduce the change in model parameter. The results in \S\ref{sec:ablation_lora} show that LoRA effectively reduces the drop in mathematical performance after training compared to full SFT. 

\subsection{RLVR with Thinking Language Selection}
The final building block is to allow on-policy thinking language selection during the traditional RLVR training process. In order to optimize the model's mathematical reasoning ability with on-policy thinking language selection, the reward function contains the following aspects:
\begin{itemize} [nosep,itemsep=2pt,leftmargin=0.2cm]
    \item Format and language compliance reward ($r_f$ and $r_c$);
    \item Thinking language Diversity reward ($r_d$);
    \item Language Pass@$k$ reward ($r_p$);
    \item Original rule-based verification reward ($r_v$);
\end{itemize}

As a result, the reward $r$ for each training sample should be 
\[r=\lambda_fr_f+\lambda_cr_c+\lambda_dr_d+\lambda_pr_p+\lambda_vr_v\]
where $r_{\cdot}$ is 0 or 1, while $\lambda_{\cdot}$ ranges from 0 to 1. Based on experimental results, the reward design of ExpLang consists of two stages, $\lambda_f=0.2, \lambda_c=0.2, \lambda_v=1$ being set for both stages, but $\lambda_d$ and $\lambda_p$ being selective to stages.

First is the \textbf{exploration stage} which covers the first 1/4 of the RL training steps. During this stage, $\lambda_d=0.2, \lambda_p=0$ is set to reward under-sampled languages to extend the exploration space. For a batch of $n$ responses, $r_d=k_{\mathrm{min}}/k$ for responses in a given language, where $k\le n$ is the response number in that language among the batch, and $k_{\mathrm{min}}$ is the smallest language response number in that batch. By such regulated but automatic thinking language selection, the model is encouraged to try different thinking languages, generating more diverse reasoning traces, while respecting answer correctness. Compared with off-policy forced thinking language selection (e.g. assigning fixed language tags during rollouts), our method is on-policy since the language selection are all acted by the model itself during training, which is acknowledged to have higher stability in RL literature \cite{andrychowiczWhatMattersOnPolicy2020}. 

Second is the \textbf{exploitation stage} which covers the remaining 3/4 of the RL training steps. During this stage, $\lambda_d=0,\lambda_p=0.5$ is set. For each correct response, all other responses in the batch with the same thinking language will be rewarded by $r_p=1$. By such design, we aim to encourage the selection of more efficient thinking languages to leverage the better-explored action space, yielding better overall performance. The model is expected to converge to one or few best-performing thinking languages. Our usage of Pass@$k$ is inspired by Chen et al.'s \yrcite{chenPasskTrainingAdaptively2025} Pass@$k$ training framework. However, instead of their idea of bootstrapping sub-groups from the rollouts, our method leverages natural sub-groups formed by thinking languages, which can also be added upon their method if needed.

The KL loss in the original GRPO algorithm is disabled during the exploration stage to enhance diversity of the model outputs, and activated during the exploitation stage to slow down the thinking language concentration.

\section{Experiment Settings}

\subsection{Models, Data and Training Strategies}
\textbf{For the models}, this study adopts the Qwen3 \cite{yangQwen3TechnicalReport2025b} model family (4B as $M_0$ for training, 32B as $\tilde{M}_0$ for data generation) for all the experiments, which is a state-of-the-art open-source LRM family with large language coverage. 
It is worth noticing that the generation length is set to 4096 for affordable training cost, instead of 32768, so the evaluated task scores will be lower than officially reported. However, this does not harm the soundness of our experiments, since all experimental inference is performed under the same length constraint, and 4096 is a meaningful length for inference with limited computational budgets.

\textbf{For the training data}, this study adopts the OpenR1-Math-220k dataset\footnote{https://huggingface.co/datasets/open-r1/OpenR1-Math-220k}, which contains high-quality competition-level math problems and ground-truth answers from NuminaMath 1.5\footnote{https://huggingface.co/datasets/AI-MO/NuminaMath-1.5} and verified thinking traces generated with DeepSeek-R1. For SFT, 7301 multilingual thinking traces in total are generated from the dataset; For RLVR, we randomly sample another 51200 samples from it.

\textbf{For the training strategies}, this study adopts LoRA \cite{huLoRALowRankAdaptation2021d} finetuning targeting all linear modules for the SFT stage base on LLaMA-Factory
\cite{zheng2024llamafactory}; And we use modified GRPO algorithm \cite{shaoDeepSeekMathPushingLimits2024} for the RLVR stage based on VeRL
\cite{sheng2024hybridflow}. These two are widely used, representative strategies among the research community. The hyper-parameters and budgets used in our training pipeline are listed in Appendix \ref{appdx:hyperparam-budget}.

\textbf{For the testing data}, this study adopts three math test sets, ranging from easy to hard: (1) MATH-500\footnote{https://huggingface.co/datasets/HuggingFaceH4/MATH-500}, which contains 500 competition math problems from the MATH benchmark \cite{hendrycksMeasuringMathematicalProblem2021,lightmanLetsVerifyStep2023a}; (2) AIME 2025\footnote{https://huggingface.co/datasets/yentinglin/aime\_2025}, which contains 30 problems from the 2025 American Invitational Mathematics Examination (AIME); (3) The en-easy subset of Olym-MATH
\cite{sun2025challengingboundariesreasoningolympiadlevel}, which contains 100  meticulously curated AIME-level problems.

\begin{table}[ht]
\caption{Performances of Qwen3 models across test sets under \textit{en} and \textit{multi} settings. ``Compl.'' denotes thinking language compliance.}
\label{tab:qwen3-potential}
\footnotesize
\centering
\begin{tabular}{@{}cccccc@{}}
\toprule
Size & Setting & Acc(\%)$\uparrow$ & Pass@k(\%)$\uparrow$ & Tokens$\downarrow$ & Compl.$\uparrow$ \\ \midrule
\multicolumn{6}{l}{\textit{MATH-500}} \\ \midrule
\multirow{2}{*}{4B} & en & \textbf{79.4} & 91.6 & 2900.7 & 100.0 \\
 & multi & 75.7 & \textbf{93.4} & \textbf{2339.7} & 73.3 \\ \cmidrule(lr){1-6}
\multirow{2}{*}{8B} & en & \textbf{77.8} & 90.6 & 3007.8 & 100.0 \\
 & multi & 75.0 & \textbf{93.4} & \textbf{2505.7} & 72.7 \\ \cmidrule(lr){1-6}
\multirow{2}{*}{32B} & en & 80.2 & 93.0 & 2796.3 & 100.0 \\
 & multi & \textbf{80.3} & \textbf{95.4} & \textbf{2242.7} & 66.9 \\ \midrule
\multicolumn{6}{l}{\textit{AIME-2025}} \\ \midrule
\multirow{2}{*}{4B} & en & \textbf{18.6} & \textbf{33.3} & 4056.2 & 100.0 \\
 & multi & 16.7 & 30.0 & \textbf{3966.7} & 71.9 \\ \cmidrule(lr){1-6}
\multirow{2}{*}{8B} & en & \textbf{21.1} & 26.7 & 4059.3 & 100.0 \\
 & multi & 14.2 & \textbf{33.3} & \textbf{3997.0} & 68.9 \\ \cmidrule(lr){1-6}
\multirow{2}{*}{32B} & en & \textbf{25.6} & 33.3 & 4011.5 & 100.0 \\
 & multi & 22.5 & \textbf{40.0} & \textbf{3858.6} & 51.7 \\ \midrule
\multicolumn{6}{l}{\textit{OlymMATH en-easy}} \\ \midrule
\multirow{2}{*}{4B} & en & \textbf{6.6} & 19.0 & 4084.9 & 100.0 \\
 & multi & 5.0 & \textbf{25.0} & \textbf{4065.7} & 73.7 \\ \cmidrule(lr){1-6}
\multirow{2}{*}{8B} & en & \textbf{5.3} & 20.0 & 4087.3 & 100.0 \\
 & multi & 4.3 & \textbf{24.0} & \textbf{4084.5} & 67.3 \\ \cmidrule(lr){1-6}
\multirow{2}{*}{32B} & en & \textbf{7.8} & 25.0 & 4084.7 & 100.0 \\
 & multi & 7.5 & \textbf{31.0} & \textbf{4040.0} & 48.7 \\ \bottomrule
\end{tabular}
\end{table}

\subsection{Baselines}
Two baselines are adopted to compare to the proposed training pipeline.
The \textbf{naive baseline} is Qwen3-4B directly trained with GRPO on the same RL training data with slightly less training cost (without LoRA) than our model, but does not suffer from catastrophic forgetting.
The \textbf{controlled baseline} is Qwen3-4B first going through LoRA SFT on controlled dataset with the same queries as for the proposed model. The dataset contains verified English thinking traces from Qwen3-32B without language selection tags. Then, the model is trained with GRPO on the RL training data. This baseline has the same training cost as the proposed model. 
The number of training samples and the hyper-parameters are consistent across the models.

\begin{table}[ht]
\caption{Performance comparison between models, with automatic (default) thinking language selection across test sets. ``Ctrl'' stands for the controlled baseline, while ``Naive'' stands for the naive baseline and ``Ours'' stands for ExpLang. Parentheses denote performance change during the last training stage.}
\label{tab:main-results-auto}
\footnotesize
\centering
\begin{tabular}{@{}lccc@{}}
\toprule
Model & Acc(\%)$\uparrow$ & Pass@k(\%)$\uparrow$ & Tokens$\downarrow$ \\ \midrule
\multicolumn{4}{l}{\textit{MATH-500}} \\ \midrule
Qwen3-4B & 78.1 & 90.0 & 2953.2 \\ \cmidrule(lr){1-4}
SFT (Ctrl) & 78.6 (+0.5) & 93.4 & 2764.1 \\
SFT (Ours) & 72.0 (-6.1) & 92.6 & 2590.9 \\ \cmidrule(lr){1-4}
RLVR (Naive) & 91.4 (+13.3) & 96.6 & 1416.9 \\
RLVR (Ctrl) & 91.0 (+12.4) & 96.6 & 1440.0 \\
RLVR (Ours) & \textbf{91.5 (+19.4)} & \textbf{96.8} & \textbf{1263.3} \\ \midrule
\multicolumn{4}{l}{\textit{AIME-2025}} \\ \midrule
Qwen3-4B & 18.9 & 30.0 & 4047.4 \\ \cmidrule(lr){1-4}
SFT (Ctrl) & 21.4 (+2.5) & 33.3 & 3996.0 \\
SFT (Ours) & 16.1 (-2.8) & 33.3 & 3937.4 \\ \cmidrule(lr){1-4}
RLVR (Naive) & 30.8 (+11.9) & 43.3 & 3408.1 \\
RLVR (Ctrl) & 26.4 (+5.0) & 43.3 & 3344.8 \\
RLVR (Ours) & \textbf{31.9 (+15.8)} & \textbf{53.3} & \textbf{3281.3} \\ \midrule
\multicolumn{4}{l}{\textit{OlymMATH en-easy}} \\ \midrule
Qwen3-4B & 7.0 & 22.0 & 4079.8 \\ \cmidrule(lr){1-4}
SFT (Ctrl) & 6.2 (-0.8) & 22.0 & 4077.7 \\
SFT (Ours) & 5.4 (-1.6) & 28.0 & 4054.9 \\ \cmidrule(lr){1-4}
RLVR (Naive) & 23.4 (+16.4) & 44.0 & 3661.3 \\
RLVR (Ctrl) & 23.2 (+17.0) & 48.0 & 3629.4 \\
RLVR (Ours) & \textbf{24.2 (+18.8)} & \textbf{53.0} & \textbf{3554.2} \\ \bottomrule
\end{tabular}
\end{table}

\section{Main Results}
\subsection{Potential Benefit of Multilingual Thinking}
We first examine the potential benefit of multilingual thinking of the Qwen3 models by measuring the models' average accuracy, Pass@$k$, language compliance and number of thinking tokens under two settings: \textit{en} (prefix-hacked English thinking for 12 randomized runs, which is close to default generation with slightly different performance), and \textit{multi} (prefix-hacked thinking in 12 non-English languages). Table \ref{tab:qwen3-potential} shows the performance of models sized 4B, 8B and 32B on the three testing datasets. One can see that, despite the highest accuracy of \textit{en}, the \textit{multi} setting shows higher Pass@$k$ values and fewer thinking tokens across model sizes and datasets. This result is in line with previous work on multilingual CoT \cite{gaoCouldThinkingMultilingually2025}. Besides, an LLM-assisted trajectory analysis shows Qwen3-32B's multilingual thinking traces form more clusters in the embedding space than English ones, indicating richer reasoning semantics (see Appendix \ref{appendix:trajectory_cluster}). These findings support our motivation that \textbf{multilingual thinking can bring potential benefit to LLM reasoning with better exploration space and more efficient reasoning.} Also, the language compliance of these models is not satisfactory, which underlines the need for a cold-start for the thinking language selection behavior.

\subsection{Performance with Automatic Selection}
In real applications, the performance with no extra constrain, i.e. under the \textit{self} setting, is the most important. Table \ref{tab:main-results-auto} shows the average accuracy, Pass@$k$ and number of thinking tokens of the proposed ExpLang method and the two baselines (naive and controlled). The results show that the proposed method exhibits the highest accuracy and Pass@$k$, indicating that \textbf{the ExpLang pipeline steadily improves the model's general reasoning ability.} 

Moreover, looking at the change in average accuracies, one can notice that our RL stage involving automatic thinking language selection brings significantly higher performance gain than the English-only baselines, suggesting that \textbf{our method can improve the efficiency of the RLVR training process within the same computational budget.} Also, it reduces thinking tokens without sacrificing performance.

\begin{table}[ht]
\caption{Results on seen languages: performance, tokens, and compliance with forced language selection. $\mathrm{Acc}^F$ means accuracy filtered on compliant samples only, and $\mathrm{Acc}^*$ means accuracy with incompliance counted as error. Other notations are consistent with previous tables.}
\label{tab:main-results-forced-seen}
\footnotesize
\centering
\setlength{\tabcolsep}{3pt} 
\begin{tabular}{@{}lcccc@{}}
\toprule
Model & $\mathrm{Acc}^F\uparrow$ & $\mathrm{Acc}^*\uparrow$ & Tokens$\downarrow$ & Compl.$\uparrow$ \\ \midrule
\multicolumn{5}{l}{\textit{MATH-500}} \\ \midrule
Qwen3-4B & 73.8 & 54.2 & 2339.7 & 73.3 \\ \cmidrule(lr){1-5}
SFT (Ctrl) & 70.5 (-3.3) & 57.3 (+3.1) & 2298.4 & 80.9 \\
SFT (Ours) & 69.0 (-4.8) & 68.6 (+14.4) & 2298.1 & 99.3 \\ \cmidrule(lr){1-5}
RLVR (Naive) & \textbf{87.2} (+13.4) & 57.8 (+3.7) & 1321.3 & 66.2 \\
RLVR (Ctrl) & 85.5 (+15.0) & 67.6 (+10.3) & 1317.8 & 78.8 \\
RLVR (Ours) & 85.7 \textbf{(+16.7)} & \textbf{85.6 (+17.0)} & \textbf{1220.0} & \textbf{99.9} \\ \midrule
\multicolumn{5}{l}{\textit{AIME-2025}} \\ \midrule
Qwen3-4B & 14.2 & 10.3 & 3966.7 & 71.9 \\ \cmidrule(lr){1-5}
SFT (Ctrl) & 14.1 (-0.1) & 11.1 (+0.8) & 3909.7 & 76.4 \\
SFT (Ours) & 14.7 (+0.5) & 14.7 (+4.4) & 3895.9 & 99.2 \\ \cmidrule(lr){1-5}
RLVR (Naive) & 23.9 (+9.7) & 15.3 (+5.0) & 3392.5 & 64.2 \\
RLVR (Ctrl) & 24.6 (+10.5) & 18.9 (+7.8) & 3332.0 & 75.3 \\
RLVR (Ours) & \textbf{27.3 (+12.6)} & \textbf{27.2 (+12.5)} & \textbf{3260.7} & \textbf{99.7} \\ \midrule
\multicolumn{5}{l}{\textit{OlymMATH en-easy}} \\ \midrule
Qwen3-4B & 3.5 & 2.6 & 4065.7 & 73.7 \\ \cmidrule(lr){1-5}
SFT (Ctrl) & 4.4 (+0.9) & 3.8 (+1.2) & 4063.5 & 84.3 \\
SFT (Ours) & 3.8 (+0.3) & 3.8 (+1.2) & 4009.1 & 99.2 \\ \cmidrule(lr){1-5}
RLVR (Naive) & 14.5 (+11.1) & 9.7 (+7.1) & 3605.7 & 67.1 \\
RLVR (Ctrl) & \textbf{16.1 (+11.7)} & 13.6 (+9.8) & 3584.3 & 82.2 \\
RLVR (Ours) & 14.7 (+10.9) & \textbf{14.7 (+10.9)} & \textbf{3566.3} & \textbf{99.7} \\ \bottomrule
\end{tabular}
\end{table}

\begin{table}[ht]
\caption{Results on unseen languages: performance, tokens, and compliance with forced language selection. The notations are consistent with previous tables.}
\label{tab:main-results-forced-ood}
\footnotesize
\centering
\setlength{\tabcolsep}{3.5pt} 
\begin{tabular}{@{}lcccc@{}}
\toprule
Model & $\mathrm{Acc}^F\uparrow$ & $\mathrm{Acc}^*\uparrow$ & Tokens$\downarrow$ & Compl.$\uparrow$ \\ \midrule
\multicolumn{5}{l}{\textit{MATH-500}} \\ \midrule
Qwen3-4B & 80.4 & 54.2 & 2201.5 & 34.0 \\ \cmidrule(lr){1-5}
SFT (Ctrl) & 70.0 (-10.4) & 32.8 (-21.4) & 2147.6 & 46.4 \\
SFT (Ours) & 64.5 (-15.9) & 41.7 (-12.5) & 2233.7 & 62.8 \\ \cmidrule(lr){1-5}
RLVR (Naive) & \textbf{83.5} (+3.1) & 24.6 (-29.6) & 1208.1 & 28.6 \\
RLVR (Ctrl) & 81.1 (+11.1) & 37.9 (+5.2) & 1258.9 & 46.6 \\
RLVR (Ours) & 81.6 \textbf{(+17.1)} & \textbf{61.5 (+19.8)} & 1222.9 & \textbf{75.5} \\ \midrule
\multicolumn{5}{l}{\textit{AIME-2025}} \\ \midrule
Qwen3-4B & 20.0 & 5.8 & 3869.0 & 33.3 \\ \cmidrule(lr){1-5}
SFT (Ctrl) & 15.0 (-5.0) & 7.5 (+1.7) & 3829.5 & 49.2 \\
SFT (Ours) & 21.1 (+1.1) & 11.7 (+5.8) & 3845.0 & 55.8 \\ \cmidrule(lr){1-5}
RLVR (Naive) & 16.7 (-3.3) & 4.2 (-1.7) & 3333.0 & 28.3 \\
RLVR (Ctrl) & \textbf{20.0} (+0.0) & 10.0 (+2.5) & 3145.0 & 46.7 \\
RLVR (Ours) & \textbf{20.0} (+0.0) & \textbf{15.0 (+3.3)} & 3249.8 & \textbf{73.3} \\ \midrule
\multicolumn{5}{l}{\textit{OlymMATH en-easy}} \\ \midrule
Qwen3-4B & 6.0 & 2.3 & 4012.1 & 34.5 \\ \cmidrule(lr){1-5}
SFT (Ctrl) & 4.5 (-1.5) & 2.3 (+0.0) & 4031.9 & 47.8 \\
SFT (Ours) & 3.3 (-2.7) & 2.5 (+0.3) & 4040.6 & 71.3 \\ \cmidrule(lr){1-5}
RLVR (Naive) & \textbf{12.5} (+6.5) & 3.3 (+1.0) & 3418.1 & 27.8 \\
RLVR (Ctrl) & 9.5 (+5.0) & 4.8 (+2.5) & 3510.2 & 45.5 \\
RLVR (Ours) & 11.0 \textbf{(+7.7)} & \textbf{8.3 (+5.8)} & \textbf{3391.6} & \textbf{74.3} \\ \bottomrule
\end{tabular}
\end{table}

\subsection{Performance with Forced Selection}
Another important setting to evaluate under is forced non-English thinking language selection, which is required to improve the readability of model thinking traces for non-English speakers. For ExpLang, this can be done by adding thinking language tags; for the baselines, this can be done by adding language-specific prefixes. Note that the language compliance of the two baselines is lower than ours, meaning they are partly using English thinking, yielding overestimated scores, so we propose two modified accuracy score: $\mathrm{Acc}^F$ that exclude incompliant samples in Acc calculation, and $\mathrm{Acc}^*$ that view incompliance as error.

As shown in Table \ref{tab:main-results-forced-seen}, in the trained 12 non-English languages, while showing comparable $\mathrm{Acc}^F$ and superior $\mathrm{Acc}^*$, our method shows higher performance gain during RLVR and fewer thinking tokens. Also, it maintains near-saturate seen thinking language compliance, much higher than the baselines. In a word, \textbf{our method significantly outperforms the baselines in terms of compliant and accurate multilingual thinking.}

Moreover, it is convenient to extend ExpLang to unseen languages, by simply providing their language selection tag. In this aspect, we take 4 unseen, low-resource languages, namely \texttt{id}, \texttt{he}, \texttt{ro} and \texttt{sw}, to evaluate the models' generalizability. Table \ref{tab:main-results-forced-ood} shows the comparison between ExpLang and the baselines in the four unseen languages. The results show that \textbf{our model substantially outperforms the baselines, while showing high unseen language compliance.} This suggests that the benefit of the ExpLang pipeline can generalize to a broader language scope.

\begin{figure}[ht]
    \centering
    \includegraphics[width=\linewidth]{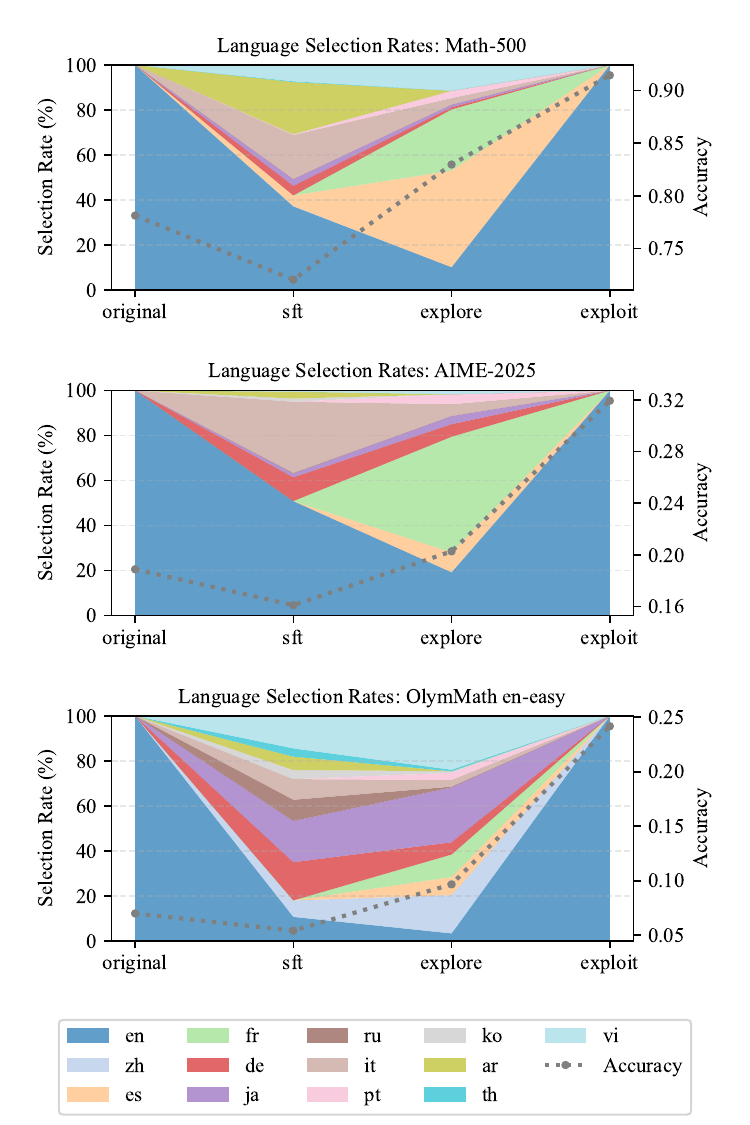}
    \vspace*{-15pt}
    \caption{Trends of thinking language selection rates of our model at the end of different training stages, across the three test sets.}
    \label{fig:selection_rates}
\end{figure}

\subsection{Changes in Thinking Language Preference}
Similar to other LRMs, the original Qwen-3 model produce all its thinking taces in English. However, during the SFT and RLVR training in our pipeline, the model's thinking language preference will go through substantial changes. Specifically, we expect the model to use non-English thinking more during and after the exploration stage, making more room of improvement for the exploitation stage.

Figure \ref{fig:selection_rates} shows the change in the model's thinking language selection rates after different training stages, which demonstrate a ``\textbf{English-Multlingual-English}'' pattern, where the accuracy under the automatic setting keeps improving during the RL stages.

After LoRA SFT stage containing many multilingual thinking samples, the model's English preference become significantly depressed. Then, after the exploration stage, English becomes equally or even less frequently chosen compared with other languages. This suggests that our SFT and RL exploration stages successfully encourage an LRM to perform multilingual thinking. 

After the exploitation stage, the model converges its thinking language selection, and English becomes dominant again. This suggests that based on the diversified feedback from the broader action space, English thinking turns out to be the most efficient in leveraging these feedback, resulting in improved overall performance. This aligns with the cross-lingual transfer of reasoning capability, where training in one language ends up improving reasoning in another language \cite{zhangGettingMoreLess2024,huLargeLanguageModels2025}, as well as the results in Table \ref{tab:main-results-forced-seen} where English-only training (``Naive'' and ``Ctrl'') also improves multilingual performance.

\section{Ablation and Analysis}

\subsection{Ablation Study}
\label{sec:ablation}
The ExpLang pipeline contains three stages: the LoRA SFT stage, the multilingual exploration stage in RL, and the language Pass@$k$ exploitation stage in RL. This section evaluates the necessity of each stage's design by comparing to the results of alternated training strategies.

\paragraph{LoRA vs. fully SFT.}
\label{sec:ablation_lora}
In the SFT stage, we adopt LoRA instead of fully finetuning to counter catastrophic forgetting which often happens to post-trained LLMs such as Qwen3. To validate our choice, we compare the LoRA-tuned model with a fully-tuned one with the same training data and hyper-parameters, except with the learning rate reduced from 1e-4 to 1e-5, which is an acknowledged operation when switching between LoRA and fully SFT \cite{bohnetComparativeAnalysisLLM2025}.

The results with automatic thinking language selection are shown in Table \ref{tab:ablate-lora}, which tells that despite being close in accuracy and language compliance under forced language selection, the LoRA model shows significantly higher accuracy than the fully finetuned one with automatic language selection. In this regard, using LoRA SFT is necessary in reducing the performance drop after training the model to follow language tags.

\begin{table}[ht]
\caption{Accuracy and language compliance of the original, LoRA-tuned and fully-tuned models. ``Auto Acc'' stands for the average accuracy with automatic thinking language selection, while ``Forced $\mathrm{Acc}^F$'' stands for the average accuracy filtered by compliance with forced thinking language selection of seen non-English languages.}
\label{tab:ablate-lora}
\centering
\footnotesize
\begin{tabular}{@{}lccc@{}}
\toprule
Model & \begin{tabular}[c]{@{}c@{}}Auto$\uparrow$\\ Acc\end{tabular} & \begin{tabular}[c]{@{}c@{}}Forced\\ $\mathrm{Acc}^F$$\uparrow$\end{tabular} & Compl.$\uparrow$ \\ \midrule
\multicolumn{4}{l}{\textit{MATH-500}} \\ \midrule
Qwen3-4B & 78.1 & 73.8 & 73.3 \\ \midrule
LoRA & 72.0 & 69.0 & 99.3 \\
Fully & \textit{66.9} & 68.7 & 99.5 \\ \midrule
\multicolumn{4}{l}{\textit{AIME-2025}} \\ \midrule
Qwen3-4B & 18.9 & 14.2 & 71.9 \\ \midrule
LoRA & 16.1 & 14.7 & 99.2 \\
Fully & \textit{10.0} & 14.4 & 99.4 \\ \midrule
\multicolumn{4}{l}{\textit{OlymMATH en-easy}} \\ \midrule
Qwen3-4B & 7.0 & 3.5 & 73.7 \\ \midrule
LoRA & 5.4 & 3.8 & 99.2 \\
Fully & \textit{3.8} & 3.4 & 99.2 \\ \bottomrule
\end{tabular}
\end{table}

\paragraph{Diversity-enhanced exploration.}
The first part of our RLVR training is the exploration stage, where we encourage the model to select less frequent languages for thinking, by which we expect to extend the exploration space and bring more room for exploitation. To validate the effect of the exploration stage, we train another model with only the exploitation stage (the total training steps matches the original exploration + exploitation steps), and compare its performance with the original one in Table \ref{tab:ablate-explore}. 

The results show that canceling the exploration step causes observable and consistent drop in performance under both the automatic and the forced settings. Since the exploitation stage also has a language compliance reward, the two models do not differ much in this aspect.

\begin{table}[ht]
\caption{Accuracy and language compliance of training with or without the multilingual exploration stage. The notations are consistent with previous tables.}

\label{tab:ablate-explore}
\centering
\footnotesize
\begin{tabular}{@{}lccc@{}}
\toprule
RL Model & \begin{tabular}[c]{@{}c@{}}Auto\\ Acc$\uparrow$\end{tabular} & \begin{tabular}[c]{@{}c@{}}Forced\\ $\mathrm{Acc}^F$$\uparrow$\end{tabular} & \begin{tabular}[c]{@{}c@{}} Compl.$\uparrow$\end{tabular} \\ \midrule
\multicolumn{4}{l}{\textit{MATH-500}} \\ \midrule
w/ explore & 91.5 & 85.7 & 99.9 \\
w/o explore & \textit{91.0} & \textit{85.0} & 99.7 \\ \midrule
\multicolumn{4}{l}{\textit{AIME-2025}} \\ \midrule
w/ explore & 31.9 & 27.3 & 99.7 \\
w/o explore & \textit{28.9} & \textit{23.9} & 100.0 \\ \midrule
\multicolumn{4}{l}{\textit{OlymMATH en-easy}} \\ \midrule
w/ explore & 24.2 & 14.7 & 99.7 \\
w/o explore & \textit{22.8} & \textit{14.1} & 99.5 \\ \bottomrule
\end{tabular}
\end{table}

\paragraph{Language Pass@k exploitation.}
The second part of our RLVR training is the exploitation stage, where we add a language sub-grouped Pass@$k$ reward to encourage the model to use its best performing languages. To validate the effect of this stage, we train another two models, one with only default GRPO reward during exploitation (``naive exploit''), the other with only the exploration stage (the total training steps matches the original exploration + exploitation steps). The results are listed in Table \ref{tab:ablate-exploit}, which offers three observations: (1) Canceling the exploitation stage causes observable drop in performance under the automatic setting, but improves performance with forced multilingual thinking, probably because the model keeps choosing different thinking languages during training and testing. (2) Naive exploitation shows close performance to multilingual Pass@$k$ exploitation, with some disadvantage in forced multilingual thinking, which shows that language Pass@$k$ training is not dominant for the improvement. (3) Although having no thinking language compliance reward during exploitation, ``naive exploit'' shows comparable compliance to the other models with the reward in exploitation, indicating that language compliance reward in either stage has a lasting effect, allowing other exploitation methods if needed.

\begin{table}[ht]
\caption{Accuracy and language compliance of training with multilingual exploitation, naive GRPO exploration, and without exploration. The notations are consistent with previous tables.}
\label{tab:ablate-exploit}
\centering
\footnotesize
\begin{tabular}{@{}lccc@{}}
\toprule
RL Model & \begin{tabular}[c]{@{}c@{}}Auto\\ Acc$\uparrow$\end{tabular} & \begin{tabular}[c]{@{}c@{}}Forced\\ $\mathrm{Acc}^F$$\uparrow$\end{tabular} & \begin{tabular}[c]{@{}c@{}} Compl.$\uparrow$\end{tabular} \\ \midrule
\multicolumn{4}{l}{\textit{MATH-500}} \\ \midrule
w/ multi. exploit & 91.5 & 85.7 & 99.9 \\
w/ naive exploit & 91.7 & 85.3 & 99.7 \\
w/o exploit & 88.0 & 87.5 & 99.9 \\ \midrule
\multicolumn{4}{l}{\textit{AIME-2025}} \\ \midrule
w/ multi. exploit & 31.9 & 27.3 & 99.7 \\
w/ naive exploit & 31.1 & 25.0 & 99.7 \\
w/o exploit & 25.6 & 29.2 & 99.4 \\ \midrule
\multicolumn{4}{l}{\textit{OlymMATH en-easy}} \\ \midrule
w/ multi. exploit & 24.2 & 14.7 & 99.7 \\
w/ naive exploit & 24.2 & 16.4 & 99.3 \\
w/o exploit & 23.8 & 17.4 & 99.8 \\ \bottomrule
\end{tabular}
\end{table}

\begin{table}[ht]
\caption{Accuracy and language compliance of training with different KL loss settings during the exploration and exploitation stages, respectively. The notations are consistent with previous tables.}
\label{tab:ablate-kl}
\centering
\footnotesize
\begin{tabular}{@{}cccc@{}}
\toprule
\begin{tabular}[c]{@{}c@{}}Stage-Wise\\ KL Use\end{tabular} & \begin{tabular}[c]{@{}c@{}}Auto\\ Acc$\uparrow$\end{tabular} & \begin{tabular}[c]{@{}c@{}}Forced\\ $\mathrm{Acc}^F$$\uparrow$\end{tabular} & Compl.$\uparrow$ \\ \midrule
\multicolumn{4}{l}{\textit{MATH-500}} \\ \midrule
(N, Y) & 91.5 & 85.7 & 99.9 \\
(N, N) & 90.8 & 85.4 & 99.7 \\
(Y, Y) & 91.1 & 84.3 & 99.8 \\ \midrule
\multicolumn{4}{l}{\textit{AIME-2025}} \\ \midrule
(N, Y) & 31.9 & 27.3 & 99.7 \\
(N, N) & 31.1 & 24.5 & 99.7 \\
(Y, Y) & 32.8 & 23.1 & 99.7 \\ \midrule
\multicolumn{4}{l}{\textit{OlymMATH en-easy}} \\ \midrule
(N, Y) & 24.2 & 14.7 & 99.7 \\
(N, N) & 24.6 & 15.8 & 99.5 \\
(Y, Y) & 23.7 & 15.0 & 99.8 \\ \bottomrule
\end{tabular}
\end{table}

\paragraph{KL loss in GRPO.}
In our RL training, we set the KL loss disabled for exploration and activated for exploitation. Here we train another two models with different settings, with the KL loss deactivated or activated all the time. The accuracy and language compliance results are shown in Table \ref{tab:ablate-kl}, where KL loss has a less significant effect on performance and language compliance. As a result, the KL loss is not a decisive factor in our training pipeline, and one may adjust the settings according to the model's use case.

\subsection{Analysis}

\begin{figure}[ht]
    \centering
    \includegraphics[width=\linewidth]{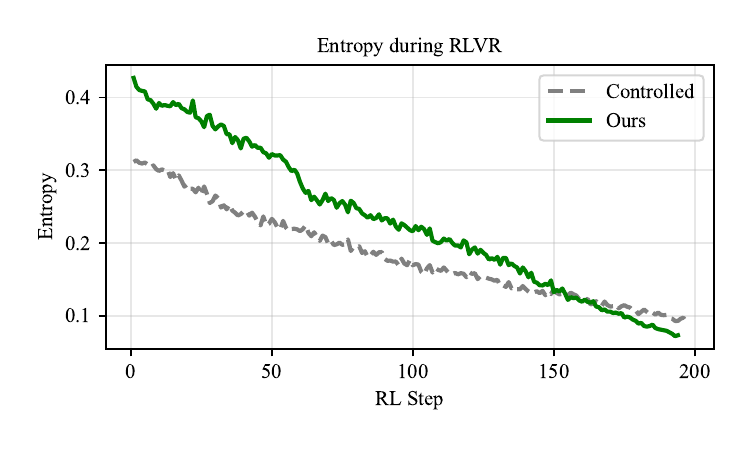}
    \vspace*{-25pt} 
    \caption{Policy entropy of ExpLang and the controlled baseline during the RLVR process.}
    \label{fig:entropy}
\end{figure}

\paragraph{Higher Response Diversity.}
The hypothesis that multilingual exploration extends the model's available action space with more diverse responses can be validated by examining the policy entropy through the training course. Figure \ref{fig:entropy} compares the entropy of our model and the controlled baseline. During exploration (step 0-49), our model shows slower decay, keeping the entropy significantly higher than the controlled baseline. During exploitation (step 50-149), our model shows faster decaying in entropy, and the final entropy becomes lower than the baseline, indicating our exploitation method helps the model to efficiently converge to an optimized policy.

\begin{figure}[ht]
    \centering
    \includegraphics[width=\linewidth]{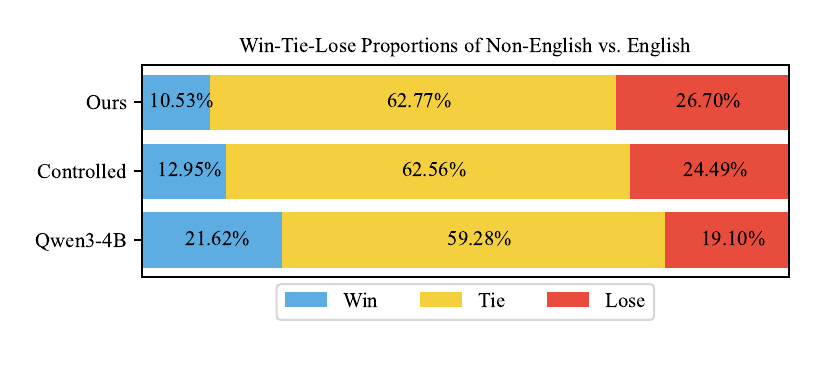}
    \vspace*{-25pt} 
    \caption{Win-Tie-Lose rate of non-English vs. English thinking, in terms of the average accuracy of the models on 2000 training samples, run for 5 times per sample.}
    \label{fig:win-tie-lose}
\end{figure}

\paragraph{Leveraged Non-English Advantage.}
Another hypothesis is that, for some training samples, the model actually performs better in certain non-English languages, but the advantage is hindered by the default English thinking mode. By enabling steady multilingual thinking, our method speeds up the learning process through leveraging the feedback of these responses. This hypothesis can be validated by examining the win-tie-lose rates in accuracy for non-English vs. English thinking on the training data. As shown in Figure \ref{fig:win-tie-lose}, for the original Qwen3-4B model, 21.62\% of the non-English thinking responses show higher accuracy than English. After training, both the controlled baseline and the proposed model show reduced non-English win rate and increased English win-rate, indicating the potential advantage of non-English thinking has been leveraged and converted into the English reasoning ability. This phenomenon is stronger for our model than for the baseline, which supports the hypothesis that our method leverages the multilingual advantage more efficiently.

\section{Conclusion and Discussion}
This paper proposes ExpLang, a novel training pipeline for LLM reasoning that enables multilingual exploration and exploitation with on-policy thinking language selection. Benefiting from higher response diversity and leveraged non-English advantage, our model shows steadily higher performance than the GRPO baselines on challenging reasoning datasets, while keeping its thinking language compliance near-perfect for seen languages and generalizable for unseen languages. Experiments show that our model goes through an ``English-Multilingual-English'' shift in thinking language preference, with the multilingual exploring stage playing the dominant role in the overall performance gain.

Beyond GRPO, the ExpLang method is orthogonal to most of the advanced RLVR algorithms and techniques for LLMs. For example, since our method only requires adding batch-computed rewards, it can be conveniently applied to other GRPO-like algorithms, such as Dr. GRPO \cite{liuUnderstandingR1ZeroLikeTraining2025a}, DAPO \cite{yuDAPOOpenSourceLLM2025b}, GSPO \cite{zhengGroupSequencePolicy2025}, etc.. However, a complete set of comparison is computational expensive, and is beyond the scope of this paper. We will actively explore such possiblity in the future.




\section*{Potential Broader Impact}
This paper presents work whose goal is to advance the field of 
Machine Learning. There are many potential societal consequences 
of our work, none which we feel must be specifically highlighted here.



\appendix

\section{Language Prefixes and Accept Rates}
\begin{figure}[ht]
    \centering
    \includegraphics[width=0.75\linewidth]{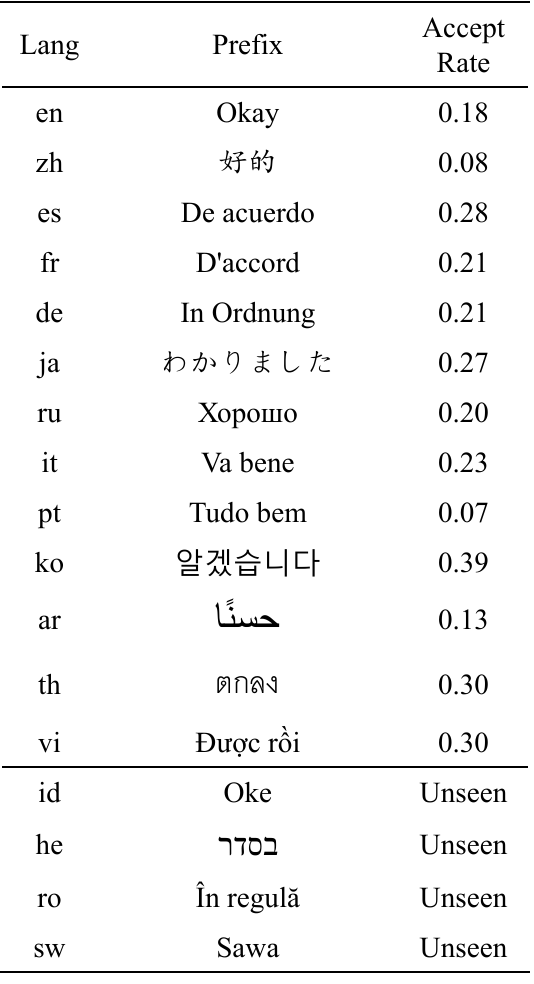}
    \caption{Prefixes and accept rates used in multilingual thinking data generation and forced multilingual thinking for baseline models.}
    \label{fig:prefixes}
\end{figure}

Figure \ref{fig:prefixes} shows the prefixes we use on the Qwen3-32B model and the baseline models to generate multilingual thinking traces. The accept rates is used only for Qwen3-32B for estimating the required inference samples so that the validated thinking traces count for about 500 per language.

\begin{figure}[ht]
    \centering
    \includegraphics[width=1\linewidth]{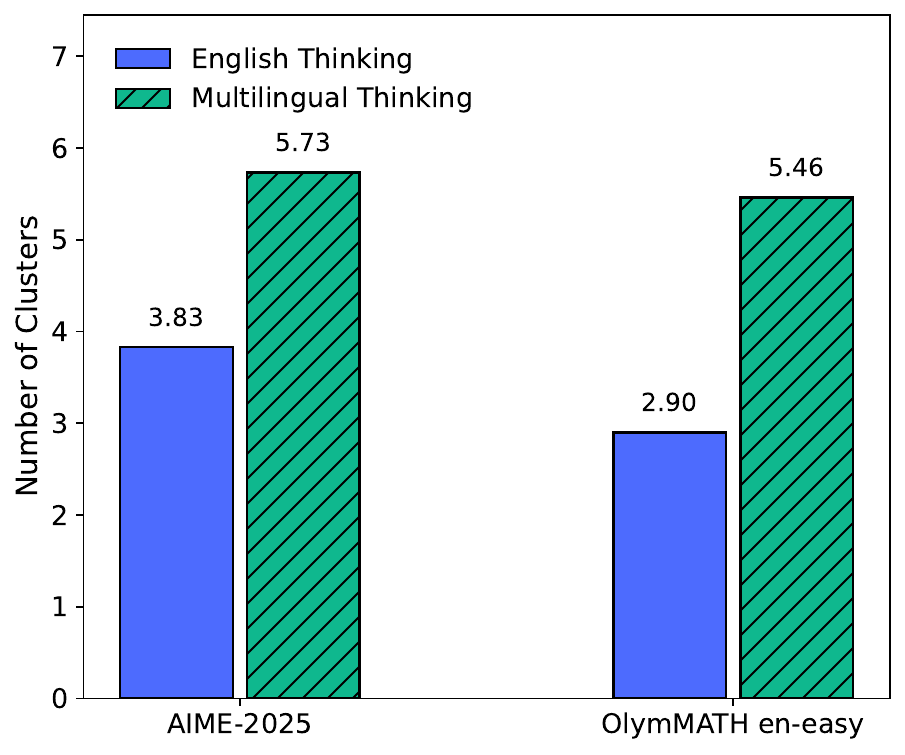}
    \caption{Comparison of Cluster Counts After Spectral Clustering of English and Multilingual Thinking.}
    \label{fig:cluster_number}
\end{figure}

\section{Trajectory Diversity of Multilingual Thinking}
\label{appendix:trajectory_cluster}
To further validate the advantage of multilingual thinking in enhancing reasoning trajectory diversity, we clustered the English and multilingual thinking outputs generated for AIME-2025 and OlymMATH en-easy, and quantified the number of resulting clusters as a measure of diversity. 
For each problem, 12 thinking samples were generated using Qwen3-32B. 
To minimize representation differences caused by multilingual expressions, we first summarize each thinking in a English text that captures its core reasoning. 
These summaries are embedded using Qwen3-Embedding-8B\footnote{\url{https://huggingface.co/Qwen/Qwen3-Embedding-8B}} and clustered using spectral clustering, and the average number of clusters per dataset is calculated.

As shown in Figure~\ref{fig:cluster_number}, multilingual thinking consistently produces more clusters than English thinking: on AIME-2025, it produces on average 1.9 more clusters and on OlymMATH en-easy, 2.56 more clusters. These results indicate that multilingual thinking has greater potential to sample a richer set of reasoning trajectories compared to English thinking.

\section{Budet and Hyper-Parameters}
\label{appdx:hyperparam-budget}
\subsection{Hyper-parameters}
\paragraph{LoRA hyper-parameters:} We set the lora target to all linear modules, rank $r=8$, $\alpha=16$, cutoff length 8192, learning rate 1e-4, with cosine scheduler and warm-up ratio 0.1. The equivalent batch size is 32 with sequence packing. For the proposed model, LoRA tuning takes 82 steps. For the controlled baseline, LoRA tuning takes 106 steps, because English thinking lengths are longer for the same queries. 

\paragraph{GRPO hyper-parameters:} We set the maximum prompt length to 2048 and the maximum response length to 4096, learning rate 1e-6, total batch size 256, mini batch size 128 (meaning updating the policy model every 2 mini-batches), 51200 training samples in total. The rollout n is 8, and the kl loss coefficient is 0.001 if applicable. For the proposed model and the baselines, the GRPO training takes 200$\pm$3 steps.

\subsection{Computational and Time Budget}
Generating high-quality multilingual thinking data with Qwen3-32B requires around 5 hours of inference with the VLLM framework\footnote{https://github.com/vllm-project/vllm} on 8$\times$Nvidia-H200 GPUs, and the corresponding English baseline data requires about 8 hours.

One run of LoRA SFT requires around 1.5 hours on 4$\times$Nvidia-A6000 GPUs, and one run of GRPO RLVR requires around 20 hours on 8$\times$Nvidia-H200 GPUs.

One run of inference on the three testing sets takes around 2.5 hours for each setting (12 random runs for each testing sample) on 4$\times$Nvidia-A6000 GPUs.


\end{document}